\documentclass[letterpaper, 10 pt, journal, twoside]{IEEEtran}

\usepackage{amssymb}  
\usepackage{graphicx} 
\usepackage{multirow}
\usepackage{multicol}
\usepackage[utf8]{inputenc}
\usepackage{amsmath,amsfonts,booktabs,cite} 
\usepackage{siunitx,textcomp} 
\usepackage[hidelinks]{hyperref} 
\usepackage{subcaption}
\usepackage[hypcap=false]{caption}

\usepackage[inline]{enumitem} 
\usepackage[nolist]{acronym} 
\usepackage{bm}
\usepackage{tikz,standalone,pgfplots}
\pgfplotsset{compat=1.13}
\pgfplotsset{ignore zero/.style={%
  #1ticklabel={\ifdim\tick pt=0pt \else\pgfmathprintnumber{\tick}\fi}
}}

\newacro{bn}[BN]{Batch Normalization}
\newacro{relu}[ReLU]{Rectified Linear Unit}
\newacro{adam}[Adam]{Adaptive Moment Estimation}
\newacro{ai}[AI]{Artificial Intelligence}
\newacro{dl}[DL]{Deep Learning}
\newacro{dnn}[DNN]{Deep Neural Network}
\newacro{cnn}[CNN]{Convolutional Neural Network}
\newacro{bnn}[BNN]{Bayesian Neural Network}
\newacro{mc}[MC]{Monte-Carlo}
\newacro{dof}[DOF]{Degrees-Of-Freedom}
\newacro{ods}[USN]{Uncertain Shape Network}
\newacro{od}[SN]{Shape Network}
\newacro{gp}[GP]{Gaussian Processes}
\newacro{gmm}[GMM]{Gaussian Mixture Model}
\newacro{gmr}[GMR]{Gaussian Mixture Regression}
\newacro{em}[EM]{Expectation Maximization}
\newacro{rl}[RL]{Reinforcement Learning}

\newacro{mcmc}[MCMC]{Markov Chain Mote Carlo}
\newacro{psdf}[p-SDF]{probabilistic Signed Distance Function}
\newacro{gpis}[GPISs]{Gaussian Process Implicit Surfaces}
\newacro{va}[V]{Varley}
\newacro{spa}[SPA]{Soft Pneumatic Actuator}
\newacro{fsr}[FSR]{Force Sensitive Resistive}
\newacro{pwm}[PWM]{Pulse Width Modulation}
\newacro{rms}[RMS]{Root Mean Square}

\newacro{pi}[PI]{Proportional-Integral}
\newacro{pid}[PID]{Proportional-Integral-Derivative}
\newacro{bic}[BIC]{Bayesian Information Criterion}
\newacro{fem}[FEM]{Finite Element Method}
\newacro{ols}[OLS]{Ordinary Least Squares}
\newcommand{\figref}[1]{\hyperref[#1]{Fig.~\ref*{#1}}}
\newcommand{\tabref}[1]{\hyperref[#1]{Table~\ref*{#1}}}
\newcommand{\secref}[1]{\hyperref[#1]{Section~\ref*{#1}}}
\newcommand{\algoref}[1]{\hyperref[#1]{Algorithm~\ref*{#1}}}

\newcommand{\ra}[1]{\renewcommand{\arraystretch}{#1}}
\newcommand{\tbs}[1]{\renewcommand{\tabcolsep}{#1pt}}

\def\methodname{DDGC}
\def\multifin{Multi-FinGAN}
\def\bestcolor{(best viewed in color)}
\def\panda{\textit{Franka Emika Panda}}
\def\barrett{\textit{Barrett hand}}
\def\sota{state-of-the-art}
\def\ie{, \textit{i.e.},}
\def\eg{, \textit{e.g.},}
\def\etal{\textit{et al.}}
\def\graspit{GraspIt!}
\def\egad{EGAD!}
\def\pc{point-cloud}
\def\pcs{point-clouds}

\def\kinect{Kinect 360\textdegree}

\def\figvspace{} 

\title{\methodname{}: Generative Deep Dexterous Grasping in Clutter}

\author{Jens~Lundell, Francesco~Verdoja, Ville~Kyrki%
\thanks{The authors are with Intelligent Robotics Group, Department of Electrical Engineering  and Automation, School of Electrical Engineering, Aalto University, 02150 Espoo, Finland (e-mail: jens.lundell@aalto.fi; francesco.verdoja@aalto.fi; ville.kyrki@aalto.fi).}%
}

\makeatletter
\let\@oldmaketitle\@maketitle
\renewcommand{\@maketitle}{\@oldmaketitle
  \setcounter{figure}{0}
  \vspace{1em}
  \includegraphics[width=\linewidth]{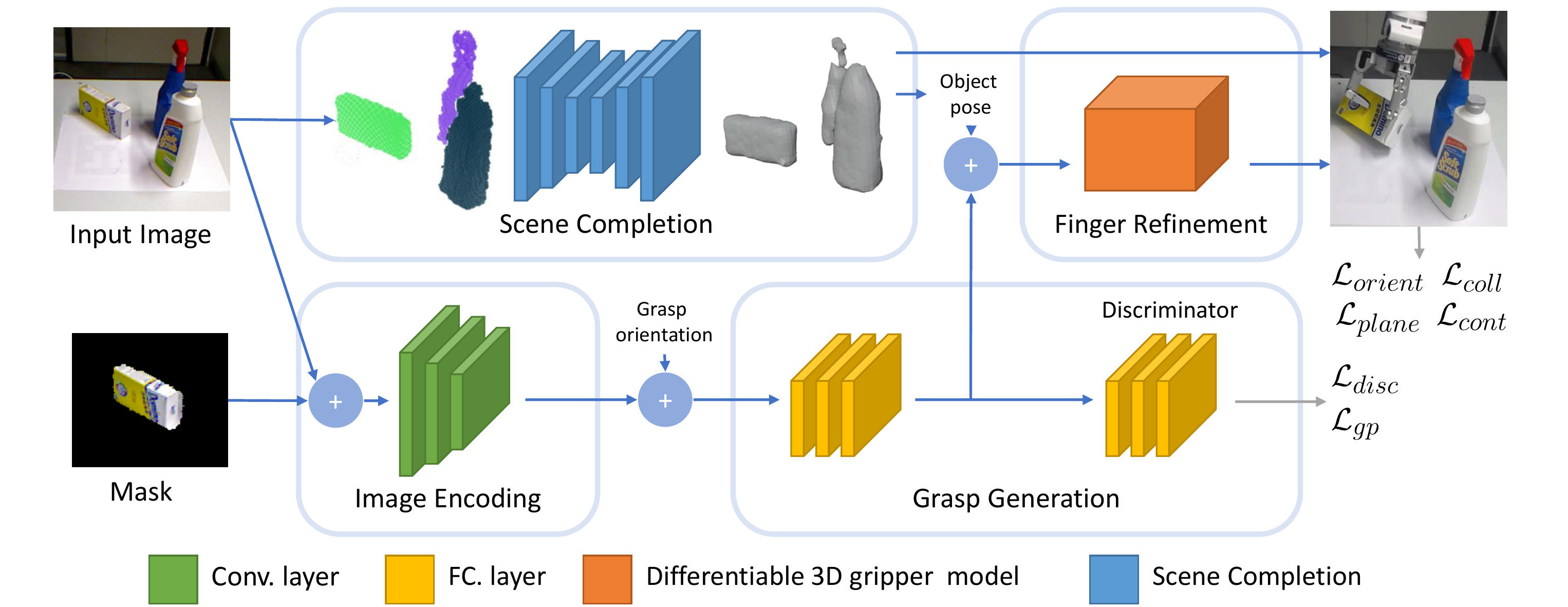}
  \captionof{figure}{\label{fig:network}Using a single RGB-D image of a cluttered scene as input, our proposed generative grasp planner can produce up to 10 collision-free multi-fingered grasps with various grasp types in less than a second. 
}}
\makeatother

\begin{document}

\maketitle

\begin{abstract}
Recent advances in multi-fingered robotic grasping have enabled fast
6-\ac{dof} single object grasping. Multi-finger grasping in cluttered scenes, on
the other hand, remains mostly unexplored due to the added difficulty of
reasoning over obstacles which greatly increases the computational time to
generate high-quality collision-free grasps. In this work, we address such limitations by introducing \methodname{}, a fast
generative multi-finger grasp sampling method that can generate high quality grasps in cluttered scenes from a single RGB-D
image. \methodname{} is built as a network that encodes scene information to produce coarse-to-fine collision-free grasp poses and configurations. We experimentally benchmark \methodname{} against two \sota{} methods on 1200 simulated cluttered scenes and 7 real-world scenes. The results show that \methodname{} outperforms the baselines in synthesizing high-quality grasps and removing clutter. \methodname{} is also 4-5 times faster than \graspit{}. This, in turn, opens the door for using multi-finger grasps in practical applications which has so far been limited due to the excessive computation time needed by other methods. Code and videos are available at \url{https://irobotics.aalto.fi/ddgc/}.
\end{abstract}


\newpage

\section{Introduction}
\label{sec:introduction}
\IEEEPARstart{R}{obotic} grasping of unknown objects in cluttered scenes, such as the one shown in \figref{fig:network}, is challenging because the scene geometry is not fully known. 
To date, the most influential results of grasp sampling in clutter are obtained with methods that sample grasps for parallel-jaw grippers~\cite{mahler2017dex, morrison2018closing, levine_learning_2016,murali20206}. 
Multi-fingered grasping introduces the added difficulty that it requires concurrent reasoning over grasp quality and collision avoidance with multi-\ac{dof} dexterous robotic hands and complex scenes. 
The benefit, however, of multi-finger grippers compared to parallel-jaw grippers is that they can attain more diverse grasp types. 
These are especially useful for task-specific grasping~\cite{kokic2017affordance}.

Current methods for multi-finger grasp generation in cluttered scenes either 
\begin{enumerate*}
    \item employ a stochastic search process such as simulated annealing to maximize a grasp quality metric to propose good grasps~\cite{miller2004graspit}, or 
    \item train a grasping policy in simulation using deep \ac{rl} to
    directly output the end pose and finger configuration of the gripper~\cite{wu2020generative}.
\end{enumerate*} 
Although a learned deep grasping policy can attain a high grasp
success rates on a diverse set of objects, it requires a simulation setup and
typically extensive parameter search to work well. Employing a stochastic search
procedure, on the other hand, does not require training any model and as
such is directly usable in a grasping pipeline; this approach is however limited
to known object models and is computationally expensive, with running times in
the order of tens of seconds to minutes.

In this work, we present \methodname{}, a deep network that can generate a set of
collision-free multi-finger grasps on unknown objects in cluttered scenes many times faster than \sota{} methods. To achieve this, we train a generative deep coarse-to-fine grasp
sampler on purely synthetic data of cluttered scenes. In comparison to similar methods where a grasping policy is trained with \ac{rl}~\cite{wu2020generative}, our method does not need any simulation setup which can be brittle to tune.  

\methodname{} was experimentally validated in simulation and in the real-world against two baselines: \multifin{}~\cite{lundell2020multi}, a recent \sota{} single-object multi-finger grasp sampling method, and the simulated-annealing planner in \graspit{}~\cite{miller2004graspit}. The simulation experiments evaluated grasp-quality and sampling speed with a \barrett{} on three different datasets containing cluttered
scenes with one to four objects. The real-world experiments, on the other hand, evaluated the clearance rate in cluttered scenes
containing 4 objects using a physical \panda{} equipped with a \barrett{}. The results show that \methodname{} generates higher
quality grasps with a higher clearance rate than both baselines. It can also generate grasps 4-5 times faster than the simulated-annealing planner in \graspit{}, which is a significant speedup.

The main contributions of this work are: 
\begin{enumerate*}[label=(\roman*)]
    \item a novel generative grasp sampling method that enables
    fast sampling of multi-finger collision-free grasps in cluttered scenes;
    \item new datasets with high-quality grasps for benchmarking multi-finger
    grasping; and
    \item an empirical evaluation of the proposed method against \sota{},
    presenting, both in simulation and on real hardware, improvements in terms
    of running time, grasp ranking, and clearance rate.
\end{enumerate*}
\section{Related work}
\label{sec:related_work}

To date, the majority of recent research on grasping in clutter has focused on parallel-jaw grippers~\cite{mahler2017dex, satish2019policy, morrison2018closing, lundell2020beyond,
levine_learning_2016, murali20206, ten2017grasp, gualtieri2016high, sundermeyer2021contact},  as the kinematic structure of such grippers simplifies the
overall problem. Only a few works~\cite{wu2020generative, varley2015generating, berenson2008grasp} consider multi-finger hands. Because of this division, we will review multi-finger grasping and grasping in clutter separately.  

\subsection{Multi-finger Grasping}

Early work on multi-finger grasping focused on constructing force-closure grasps
that are dexterous, in equilibrium, stable, and exhibit a certain dynamic
behavior by casting the problem as an optimization problem over one or all of
these criteria~\cite{sahbani2012overview, berenson2008grasp}. However, one of
the core issues of such methods are that they assume precise and known geometric
and physical models of the objects to grasp~\cite{bohg_data-driven_2014}.
Moreover, analytical methods are typically computationally expensive and as such
generate only a few grasps. 

To lift some of the requirements of analytical methods and to make grasp
sampling faster, researchers resorted to data-driven methods. Early data-driven
methods are based heavily on the development and release of \graspit{} in 2004~\cite{miller2004graspit} and focused mainly on the issue of sampling grasps from
the infinite space of solutions~\cite{miller_automatic_2003, ciocarlie2009hand,
borst2003grasping, goldfeder2007grasp, pelossof2004svm}. Out of these works, the
\textit{eigengrasp} formulation by Ciocarlie and Allen~\cite{ciocarlie2009hand}---where grasp planning is performed in a hand posture
subspace of highly reduced dimensionality---has proven effective as it is still
used today as a grasp sampling component in many grasping pipelines~\cite{varley_shape_2017, lundell2019robust, watkins-valls_multi-modal_2018}.
Although these methods greatly sped up grasp sampling in comparison to
analytical methods, they still operate in the order of tens of
seconds to minutes and require full object models.   

To enable data-driven grasp sampling methods to work on unknown objects, recent
works have used deep learning to shape complete unknown objects and then plan
grasps on the completed shapes~\cite{varley_shape_2017, agnew2020amodal,
lundell2019robust, watkins-valls_multi-modal_2018}. These methods generalize
across many different objects but still rely on an inherently slow sampling
process. Therefore, other deep-learning methods have focused on learning the
actual grasp sampler either from raw sensor inputs~\cite{wu2020generative,
shao2020unigrasp, kokic2017affordance, aktas2019deep, lu2020active} or shape
completed models~\cite{lundell2020multi}. The limitation of all of these
methods, with the exception of~\cite{wu2020generative}, is that they are
proposed for and evaluated on single object grasping. In comparison, the method
presented here can generate diverse high-quality grasps fast on both single
objects and objects in clutter. 

\subsection{Grasping in Clutter}

Grasping in clutter poses additional constraints on grasp sampling as it
typically requires grasps to not collide with obstacles. Therefore, to simplify
the grasp planning problem in clutter much work resorts to planning grasps with
simpler parallel-jaw grippers opposed to more complex dexterous hands~\cite{mahler2017dex, satish2019policy, morrison2018closing, lundell2020beyond,
levine_learning_2016, murali20206, ten2017grasp, gualtieri2016high, sundermeyer2021contact}. Murali
\etal{} \cite{murali20206} split the concept of grasping in clutter into the following
subcategories: bin-picking vs. structured clutter, planar vs. spatial grasping
in clutter, model-based vs. model-free, and target-agnostic vs. target-driven. 

In bin-picking, objects are located in a pile and are often small and light.
Methods that target bin-picking~\cite{mahler2017dex, satish2019policy,
morrison2018closing, lundell2020beyond, levine_learning_2016} benefit from the
pile structure as objects have more stable equilibriuma and as such collisions
with obstacles do not jeopardize the pile structure. In comparison, methods
that target grasping in structured clutter~\cite{murali20206, sundermeyer2021contact}---where objects
are mostly larger, heavier, and packed together---need to avoid all collisions
as unintended contact between the robot and objects can easily tip objects over
and thus change the scene structure. Our approach calculates if a grasp is in
collisions by checking if the hand mesh is colliding, while~\cite{murali20206}
requires another trained network to predict potential collisions.

In cluttered scenes, the most successful methods have resorted to the simpler
4-\ac{dof} planar top-down grasps~\cite{mahler2017dex, satish2019policy,
morrison2018closing, levine_learning_2016} instead of full 6-\ac{dof} spatial
grasps~\cite{murali20206, wu2020generative, lundell2019robust, sundermeyer2021contact}. While 4-\ac{dof}
grasps are simpler, the restriction on the arm motion they imply hinders the
robot's ability to grasp specific objects~\cite{wu2020generative, murali20206, sundermeyer2021contact},
especially with structured clutter. Because of these reasons, we developed our
method to predict full 6-\ac{dof} grasps.

Model-based grasping in clutter assumes known object models and as such
simplifies the planning problem~\cite{berenson2008grasp}. Approaches that target
model-free grasping in clutter where objects are unknown either plan grasps
directly on raw sensor inputs~\cite{mahler2017dex, satish2019policy,
morrison2018closing, levine_learning_2016, murali20206, wu2020generative,sundermeyer2021contact} or
first estimate the shape of the unknown objects and then plan grasps~\cite{lundell2020beyond, varley2015generating}. In this work, we estimate the
shape of all the objects, as it allows us to plan grasps that are closer to the
surface of the objects.

With target-driven grasp sampling methods~\cite{murali20206} it is possible to
specify the object to grasp opposed to target-free methods~\cite{mahler2017dex,
satish2019policy, morrison2018closing, lundell2020beyond, levine_learning_2016,
wu2020generative,sundermeyer2021contact} which oftentimes simply chooses the object it can
attain the best grasp on. Our method is target-driven as it segments objects and
plan object-specific grasps using these segments.     

In a nutshell, the work presented here falls into the category of target-driven
model-free multi-finger spatial grasping in scenes with structured clutter. A similar multi-finger grasping in clutter approach was proposed in~\cite{wu2020generative} but for a different use-case: target-agnostic bin-picking. That method used \ac{rl} to train a deep grasping policy in simulation which transferred seamlessly to
the real world, attaining a high grasp success rate on a diverse set of
objects. Our grasp sampler is also trained on purely synthetic data but, in contrast to \cite{wu2020generative}, does not require an extensive simulation setup to train.

\methodname{} goes beyond prior work on grasping in clutter with parallel-jaw
grippers~\cite{mahler2017dex, satish2019policy, morrison2018closing,
lundell2020beyond, levine_learning_2016, murali20206, ten2017grasp,
gualtieri2016high,sundermeyer2021contact} by using a more complex dexterous gripper and reasoning over
more \ac{dof}, generalizing to a more complicated problem. In comparison to
previous work on grasping in clutter with dexterous hands~\cite{wu2020generative, berenson2008grasp, varley2015generating}, our work goes
beyond those by either being considerably faster (8 sec. vs 16 sec \cite{varley2015generating}) or allowing to select the object to grasp \cite{wu2020generative,berenson2008grasp}.
\section{Problem formulation}
\label{sec:prob_form}

In this work, we address the problem of grasping unknown objects in structured
clutter with a multi-finger robotic gripper. We realize this by predicting a 6D
gripper pose $\mathbf{p}$ and a hand joint configuration $\mathbf{q}$ for which
the grasp does not collide with any obstacle but still has several contact
points with the target object to ensure a robust grasp. More formally, we train
a parametric model $\mathcal{M}_{\bm{\theta}}$ with parameters $\bm{\theta}$
that takes as input an RGB-D image $\mathbf{I}$ and produces a 6D gripper pose
$\mathbf{p}$, and a hand joint configuration $\mathbf{q}$:

\begin{equation*}
    \mathcal{M}_{\bm{\theta}}: \mathbf{I} \to \left( \mathbf{p}, \mathbf{q} 
    \right) \enspace.
\end{equation*}
The poses $\mathbf{p}$ of all grasps are in the target object's center of reference. 
 
We further constrain the target hand joint configuration $\mathbf{q}$ to leave a small
clearance between the fingers and the target object, and the gripper is closed until contact before attempting to lift an object. This
is done to limit the impact of sensing uncertainties in the
real-world as we avoid the need to generate precise configurations that 
touch the surface of the object.
\section{Method}
\label{sec:method}

To generate high-quality collision free grasps in cluttered scenes, we introduce
the model shown in \figref{fig:network} which is inspired from prior work on
human-hand grasp generation~\cite{corona2020ganhand} and our own work on
predicting multi-finger grasps on single objects~\cite{lundell2020multi}. This
network consists of sub-modules that were used for single object grasping along
with novel ones that were essential to adapt the architecture to grasping in
clutter. Therefore, for completeness, we will next briefly overview the entire
model and then describe the novel technical contributions in detail.

\subsection{Network Overview}
\label{sec:network_overview}
The network in \figref{fig:network} is split into 5 sub-modules: Scene
Completion, Image Encoding, Grasp Generation, Discriminator and Finger
Refinement. Out of these Image Encoding and Grasp Generation are novel while the
others have been used in prior work~\cite{corona2020ganhand, lundell2020beyond,
lundell2020multi}.

\emph{Scene Completion} refers to the process of producing a full 3-D model of the scene by segmenting the scene and shape completing
each object~\cite{lundell2020beyond}. The method is agnostic to the
segmentation algorithm. For shape completion, we use a pre-trained fully
convolutional hour-glass shaped \ac{dnn} that operates on voxelized point-clouds
as input and outputs a completed voxel grid of the object. The object is
post-processed into a mesh by first merging it with the input point-cloud and
then running the marching cube algorithm~\cite{lorensen_marching_1987}. The
shape completed objects are later used for grasp generation, finger refinement, and collision detection.

The \emph{Image Encoding} module, further detailed together with the Grasp
Generation module in the next section, produces an encoding $\text{E}(\mathbf{I})$ of the
geometry of the scene from an RGB image $\mathbf{I}$ using a \ac{cnn}. This encoding, together with an input
grasp orientation $\textbf{r}_0$, is then passed through the fully-connected \emph{Grasp
Generation} network to generate a coarse gripper joint configuration
$\mathbf{q}_c$ and a 6D coarse object-centric grasp pose $\mathbf{p}_c=(\mathbf{t}_c,\mathbf{r}_c)$ where
$\mathbf{t}_c$ represents a translation vector and $\mathbf{r}_c$ rotation as normalized axis angles. The coarse grasp pose is then refined into global coordinates $\mathbf{p}^*=(\mathbf{t}_c+ \mathbf{t}_0,\mathbf{r}_c+\mathbf{r}_0)=(\mathbf{t},\mathbf{r})$ by adding the center of the mesh $\mathbf{t}_0$ to the coarse translation vector $\mathbf{t}_c$ and the input rotation $\mathbf{r}_0$ to the coarse rotation $\mathbf{r}_c$. 

When training \methodname{}, the input
rotation $\textbf{r}_0$ is set to a grasp rotation from the dataset with added zero-mean Gaussian noise while, at test-time, the input rotations are randomly sampled from around the object. Similarly, the center of the
target object $\mathbf{t}_0$ is known during training because we are operating on known meshes,
while during the physical experiments we use the center of the shape-completed
mesh.

In this work, the grasp configuration $\mathbf{q}_c$ represents the 7-\ac{dof} of a Barrett hand shown in \figref{fig:successful_and_failed_grasps}. However, the Grasp Generation network
is only tasked to output the finger spread of the coarse gripper configuration
$\mathbf{q}_c$ and leaves the rest of the \ac{dof} in a default zero position.
The reason for doing this is that we use the Finger Refinement layer to 
refine the coarse gripper joint configuration $\mathbf{q}_c$ to be close to the
target object. 

The \emph{Finger Refinement} layer is tasked to rotate each finger of the gripper until it touches the object. It is represented as the forward kinematics of
the gripper and as such is fully differentiable and does not include any
additional trainable parameters. Given an input 6D pose $\mathbf{p}^*$ and coarse
hand configuration $\mathbf{q}_c$ it produces a refined configuration
$\mathbf{q}^*=\mathbf{q}_c+\Delta\mathbf{q}$ by rotating each articulated finger
$\Delta\mathbf{q}$ radians until the distance between the finger and the object
mesh is below a predefined threshold $t_d$ which we set to 1 cm.

To ensure that the distribution of generated grasps is close to the training
distribution we add a Wasserstein discriminator D~\cite{martin2017wasserstein}
trained with the gradient penalty~\cite{gulrajani2017improved}. Given the grasp
generator G, the objective function to minimize is

\begin{align}
\begin{split}
    \mathcal{L}_{disc} =& \mathop{\mathbb{E}}\left[\text{D}(\text{G}(\text{E}(\mathbf{I}),\mathbf{r}_0))\right]-\mathop{\mathbb{E}}\left[\text{D}(\widehat{\mathbf{q}},\widehat{\mathbf{t}},\widehat{\mathbf{r}})\right]\enspace,\\
    \mathcal{L}_{gp} =& \mathop{\mathbb{E}}\left[\left(\|\nabla_{\widetilde{\mathbf{q}},\widetilde{\mathbf{T}},\widetilde{\mathbf{r}}} \text{D}(\widetilde{\mathbf{q}},\widetilde{\mathbf{t}},\widetilde{\mathbf{r}}) \|_2 -1\right)^2\right]\enspace,
\end{split}
\end{align}
where $\widehat{\mathbf{q}}$, $\widehat{\mathbf{t}}$, and $\widehat{\mathbf{r}}$
are samples from the training dataset and $\widetilde{\mathbf{q}}$,
$\widetilde{\mathbf{t}}$, and $\widetilde{\mathbf{r}}$ are linear interpolations
between predictions and those samples.

\subsection{Multi-finger Deep Grasping in Clutter}

As mentioned in the previous section the novel sub-modules in this work are the
\emph{Image Encoding} and \emph{Grasp Generation} networks. The use of these two
networks were, in part, explored in our previous work on single object robotic
grasping~\cite{lundell2020multi}. However, the architectures we proposed there
cannot handle the complexity of cluttered scenes because of the following
reasons:
\begin{enumerate*}
    \item the image encoding network operates on input RGB images where
    everything except the target object is masked out,
    \item the grasp generation network requires a coarse estimate of the input
    grasp configuration.
\end{enumerate*}

In a cluttered scene, a consequence of encoding only image information about the
target object is that information about obstacles is ignored and as such cannot
prime the grasp sampler to generate collision-free grasps. To circumvent this
issue, we kept the original input RGB image unmasked and instead added an
additional channel with the target object mask. When training the network we
have ground-truth masks of all objects, while at run-time we produce the masks from a segmented \pc{} of the scene. An additional benefit of adding the mask of the
object is that we effectively make the complete grasp sampling network
target-driven, that is we can choose which object to grasp by selecting the
corresponding mask.

In our previous work on dexterous multi-finger grasping~\cite{lundell2020multi},
the grasp generation network also produced residual grasp joints that were added
to initial ones given from a classification network. The classification network
classified grasps according to one of the 33 grasp taxonomies listed in~\cite{feix2015grasp}. However, this required both to train an additional network
and to label all grasps according to their taxonomy which is 
time-consuming as it requires human domain knowledge to correctly label the
grasps.

\subsection{Loss functions}

An empirical finding in our previous work on multi-finger grasp sampling of
single objects~\cite{lundell2020multi} is that the discriminator helps in
producing realistic-looking grasps but it alone cannot guide the training
enough. Therefore, we add the following complementary loss functions: a contact
loss $\mathcal{L}_{cont}$, a collision loss $\mathcal{L}_{coll}$, an orientation
loss $\mathcal{L}_{orient}$, and a plane loss $\mathcal{L}_{plane}$. Out of
these  $\mathcal{L}_{cont}$, $\mathcal{L}_{coll}$, and $\mathcal{L}_{orient}$
were used in our previous work~\cite{lundell2020multi}, while $\mathcal{L}_{plane}$ was originally
introduced in~\cite{corona2020ganhand}.

For grasps to be successful they need to have multiple contact points with the
object. To encourage such behavior we added the contact loss defined as 
\begin{align}
\mathcal{L}_{cont} = \frac{1}{|V_{cont}|}\sum_{v\in V_{cont}}\min_{k}\|v,O_k\|_2\enspace,
\end{align}
where $O_k$ are the object vertices. $V_{cont}$ is the subset of vertices on the hand that
are often in contact with the target object\ie{} vertices that were closer than 5 mm to the object in at least 8\% of the grasps in the dataset. Vertices in $V_{cont}$ are mainly located on the gripper's fingertips and palm.

In our previous work~\cite{lundell2020multi}, we identified having a loss that
encourages grasps to point towards the object to grasp as helping the training.
This is realized by encouraging the approach vector of the gripper
$\mathbf{\hat{a}}$ to point in the same direction as the vector
$\mathbf{\hat{o}}$ connecting the hand to the object's center:  
\begin{align}
\mathcal{L}_{orient} = 1 - \mathbf{\hat{a}}^\top\mathbf{\hat{o}}\enspace.
\end{align}

To penalize collisions between grasps and the environment which, in scenes with
structured clutter is of uttermost importance, we added the following loss
\begin{align}
\label{eq:coll_loss}
\mathcal{L}_{coll} = \frac{1}{|V_i|}\sum_{j}^{|O|} \sum_{\mathbf{v}\in V_i} \text{A}_\mathbf{v}\min_k \|\mathbf{v},O_k^j\|_2\enspace.
\end{align}
This loss penalizes the distance between all vertices of a grasp that lie inside
an object $\mathbf{v}\in V_i$ and their closest surface points to object
$O^j$, where $|O|$ is the number of objects present in the scene. We generate
surface points of every object by uniformly sampling $k$ points from their mesh.
The weight $A_v$ is the average area of all incident faces to a vertex $v$, which is important when working with non-uniform gripper tessellations~\cite{lundell2020multi}.

Finally, we also need to encourage grasps to stay above the plane the objects are
resting on. By representing the plane as a point $\mathbf{p}_t$ on its surface
and its normal $\mathbf{\hat{n}}_t$, we can penalize all vertices $\mathbf{v}$
on the gripper mesh that are located under the table with    
\begin{align}
\mathcal{L}_{plane} = \sum_{\mathbf{v}\in V} \min \left(0,(\mathbf{v}-\mathbf{p}_t)^{\text{T}}\cdot\mathbf{\hat{n}}_t \right).
\label{eq:plane_loss}
\end{align}
Equation \eqref{eq:plane_loss} is only negative when the normal vector $\mathbf{\hat{n}}_t$ and the
vector connecting the point on the plane $\mathbf{p}_t$ and a vertex
$\mathbf{v}$ points in different directions indicating that the vertex
$\mathbf{v}$ is located below the table.

The final loss is a weighted linear combination of the six individual losses
$\mathcal{L}=w_{disc}\mathcal{L}_{disc}+w_{gp}\mathcal{L}_{gp}+w_{cont}\mathcal{L}_{cont}+w_{coll}\mathcal{L}_{coll}+w_{orient}\mathcal{L}_{orient}+w_{plane}\mathcal{L}_{plane}$.
The network is trained end-to-end.

\subsection{Implementation details}

\begin{figure}%
	\centering
	\begin{subfigure}[b]{0.24\linewidth}
		\centering
		\includegraphics[width=\linewidth]{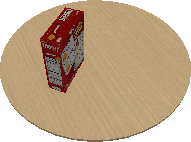}%
	\end{subfigure}
	\begin{subfigure}[b]{0.24\linewidth}
		\includegraphics[width=\linewidth]{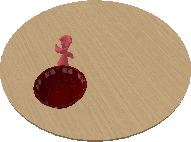}%
	\end{subfigure}
	\begin{subfigure}[b]{0.24\linewidth}
		\includegraphics[width=\linewidth]{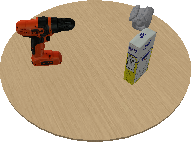}%
	\end{subfigure}
	\begin{subfigure}[b]{0.24\linewidth}
		\includegraphics[width=\linewidth]{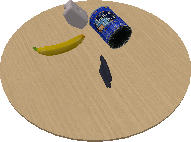}%
	\end{subfigure}
	\caption{\label{fig:examples}Example scenes from our data-set with one up to four objects.}
	\figvspace{}
\end{figure}

The network was implemented in PyTorch 1.5.1. The model was trained
on scenes containing 1, 2, 3, or 4 objects randomly placed on a table surface;
some examples are shown in~\figref{fig:examples} and details on how we generated them are given in \secref{sec:dataset}. The images were resized to
$256 \times 256$. We trained our networks with a learning rate of $10^{-4}$ and a
batch size of 64. The image encoder was fine-tuned on weights from a pre-trained ResNet-50 while the weights for the shape completion network were not trained at all and instead set to the same as in \cite{lundell2019robust}. The generator was trained once every 5 forward passes to improve the relative
quality of the discriminator. We trained the networks for 12000 epochs, reducing the learning rate linearly for the last 4000 epochs.

The weights of the loss functions were set to the same values as in \cite{lundell2020multi} except the plane loss which was set to the same value as reported in \cite{corona2020ganhand}. During training, we did not witness any detrimental competition between the losses. Instead, they reduced in tandem.
\section{\methodname{} dataset}
\label{sec:dataset}

Training deep networks, such as \methodname{}, typically requires a lot of
diverse training data to obtain good performance. Moreover, the usual assumption
is that train and test data should originate from a similar underlying
distribution. This, in the case of real-world robotics experiments, means we
need to gather training data from real robot interactions, a procedure that is
both time-intensive and wears the robotic hardware. Many prior works have,
however, empirically shown that training grasp generation networks on simulated
data alone opposed to on real-world data works well even for
real-world grasping~\cite{morrison2018closing, mousavian20196,
lundell2020multi}.

Based on these findings, we train \methodname{} on only simulated data of
cluttered scenes. To simulate cluttered scenes we use the physics simulator
PyBullet~\cite{coumans2019}. A scene is generated by first selecting the number
of objects to place in it and then, for one object at a time, we randomly
generate a position on the table that is not in collision with any other object
and place it there in one of its stable resting poses\footnote{The stable resting poses were calculated using
trimesh (\href{https://github.com/mikedh/trimesh}{github.com/mikedh/trimesh})}. If objects have
predefined textures we apply them, otherwise, we randomly generate texture maps.
Finally, we render the RGB image and the object masks from a 45\si{\degree}
viewing angle and store them together with the object poses. 

Training \methodname{} requires not only RGB and masked images but also
multi-finger grasps on each object. Multi-finger grasps were first
generated separately for each object in their canonical pose using \graspit{}.
To avoid bias and to generate a dense set of grasps we uniformly sampled grasps around the object
with two different distances to the object (2 and 8 cm) and three
different finger spreads (0\si{\degree}, 45\si{\degree}, and 90\si{\degree}). These object-centric grasps were then transferred to the
cluttered scenes by applying the same random transform that the objects went
through when placing them in the scene. A grasp was marked successful if it was
collision-free.

We generated one training set of cluttered scenes on 18 objects from the YCB
object set~\cite{calli2015ycb} and 120 objects from the \egad{} training set~\cite{morrison2020egad}. We also generated three validation sets: one on the 49
validation objects in the \egad{} validation set, one on the 152 objects in the
KIT object model database~\cite{kasper2012kit}, and one with both \egad{}
validation set and KIT object model database. To generate datasets with a
varying amount of clutter, we generated 100 random scenes with 1, 2, 3 or 4
objects in it, totaling 400 scenes per dataset.
\section{{Experiments and Results}}
\label{sec:exp_and_res}

The two main questions we wanted to answer in the experiments were:
\begin{enumerate}
    \item Is \methodname{} able to generate high quality grasps in scenes with
    clutter?
    \item Is our generative grasp sampler, which is purely trained on synthetic
    data, able to transfer to real objects?
\end{enumerate}

In order to provide justified answers to these questions, we conducted two
separate experiments. In the first experiment (\secref{exp:grasp_in_sim}) we
evaluate grasp quality and hand-object interpenetration in simulation. In the
second experiment (\secref{exp:physical_grasping}) we evaluate the clearance rate
in cluttered scenes on real hardware.

In all our experiments, we benchmark our approach against \multifin{} \cite{lundell2020multi} and the simulated-annealing
planner in \graspit{}~\cite{miller2004graspit} which is the only planner that can do multi-finger target-driven grasping in clutter. We let the simulated-annealing
planner run for 75000 steps. To run \graspit{} with clutter, the target object
was set as a graspable object while the rest of the objects were set as
obstacles. We also included the table the objects were resting on as an
obstacle.  

In both experiments, \methodname{} and \multifin{} generated 100 grasps per object. Grasps that were in collision with the table or any other objects were removed. As we a-priori knew that all objects rest on a flat surface we only sampled, for both \methodname{} and \multifin{}, input rotations
$\mathbf{r}_0$ that rotated the gripper's approach direction to point towards
the negative normal direction of the supporting surface. This heuristic
effectively removed input rotations that correspond to grasps coming from below the table. 

\subsection{Grasping in Simulation}
\label{exp:grasp_in_sim}
\begin{table*}
\centering
\ra{1.3}\tbs{6}
\caption{\label{tb:sim_exp_summary}Simulation experiment results. $\uparrow$: higher the better; $\downarrow$: lower the better.}
\begin{tabular}{lccccccccc}
\toprule
 & \multicolumn{3}{c}{\graspit} & \multicolumn{3}{c}{Multi-FinGAN} & \multicolumn{3}{c}{\methodname} \\
\cmidrule(lr){2-4} \cmidrule(lr){5-7} \cmidrule(lr){8-10}& \begin{tabular}[c]{@{}c@{}}\egad{} \\ val\end{tabular} & KIT & \begin{tabular}[c]{@{}c@{}}\egad{} val \\ + KIT\end{tabular} & \begin{tabular}[c]{@{}c@{}}\egad{} \\ val\end{tabular} & KIT & \begin{tabular}[c]{@{}c@{}}\egad{} val \\ + KIT\end{tabular} & \begin{tabular}[c]{@{}c@{}}\egad{} \\ val\end{tabular} & KIT & \begin{tabular}[c]{@{}c@{}}\egad{} val \\ + KIT\end{tabular} \\  \midrule
Avg. $\epsilon$-quality over 10 grasps $\uparrow$ & 0.21 & 0.20 & 0.20 & 0.32 & 0.38 & 0.35 & \textbf{0.74} & \textbf{0.73} & \textbf{0.72} \\
Avg. interpenetration over 10 grasps ($\text{cm}^3$) $\downarrow$ & \textbf{2.92} & 4.76 & \textbf{3.63} & 4.12 & 5.92 & 5.05 & 4.00 & \textbf{3.30} & 3.64 \\
Grasp Sampling for 10 grasps (sec.) $\downarrow$ & 44.09 & 39.97 & 35.75 & \textbf{9.10} & \textbf{8.80} & \textbf{9.00} & 9.60 & 9.40 & 9.40 \\ 
\bottomrule
\end{tabular}
\vspace{-1em}
\end{table*}

\begin{figure}
	\centering
	\begin{subfigure}[b]{.23\textwidth}
		\centering
            \begin{tikzpicture}[scale=0.5]
    \begin{axis}[
    	xlabel={\Large Number of objects in scene},
    	ylabel={\Large $\epsilon$-quality},
        xtick distance=1,
        legend style={at={(1,0.7)},anchor=east},
        tick label style={font=\Large}
        ]
    
    \addplot[color=red,mark=x] coordinates {
    	(1, 0.39)
    	(2, 0.19)
    	(3, 0.13)
    	(4, 0.1)
    };
        \addplot[color=black,mark=o] coordinates {
    	(1, 0.34)
    	(2, 0.35)
    	(3, 0.37)
    	(4, 0.35)
    };

    \addplot[color=blue,mark=*] coordinates {
    	(1, 0.73)
    	(2, 0.72)
    	(3, 0.72)
    	(4, 0.72)
    };
    \legend{\graspit,\multifin, \methodname}
    \end{axis}
    \end{tikzpicture}
        \caption{\label{fig:epsilon_quality_over_clutter}}
	\end{subfigure}
	\begin{subfigure}[b]{.24\textwidth}
        \centering
            \begin{tikzpicture}[scale=0.5]
    \begin{axis}[
        ymin=0,
        minor y tick num = 1,
        bar shift=0pt,
        area style,
        xlabel={\Large Spread angle (\textdegree)},
        ylabel={\Large Fraction of grasps},
        axis x line=bottom,
        axis y line=left,
        xtick distance=18,
        ytick distance=0.1,
        xmin=0,xmax=90,
        tick label style={font=\Large},
        ignore zero=y
        ]
    \addplot+[ybar interval,mark=no] plot coordinates {(0,0.54) (18,0.34) (36,0.26) (54,0.2) (72,0.33) (90,0)};
    \end{axis}
    \end{tikzpicture}
        \caption{\label{fig:finger_spread_histogram}}
    \end{subfigure}
	\caption{\label{fig:behaviour_over_clutter}(a) shows that \methodname{} constantly finds higher-quality grasps than both \graspit{} and \multifin{}. (b) shows a histograms of the finger spread of grasps generated with \methodname.}
	\figvspace{}
\end{figure}
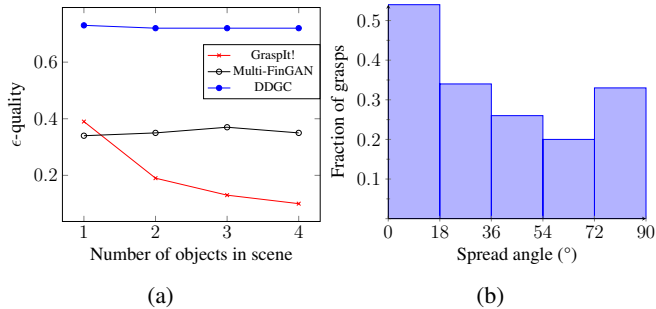

Here we present quantitative results of grasps generated by  \methodname{}, \multifin{}, and the
simulated-annealing planner in \graspit{} on the three
different held out validation sets: \egad{} val, KIT, and \egad{} val + KIT. To
quantify the quality of a grasp we calculated both its intersection
with all objects and the well known $\epsilon$-quality metric~\cite{miller1999examples} which represents the radius of the largest 6D ball
centered at the origin that can be enclosed by the wrench space's convex hull. 

For all methods, we rank the grasps according to their
$\epsilon$-quality and report the average performance on the 10 top-scoring
grasps. We also report how the metrics vary across increasing amount of clutter.

\begin{figure}
	\centering
		\begin{subfigure}[b]{.15\textwidth}
		\centering
		\includegraphics[width=\linewidth]{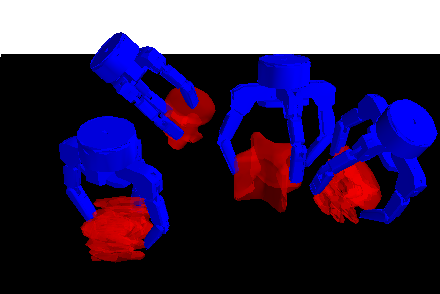}%
	\end{subfigure}\hspace{-0.3em}
	\begin{subfigure}[b]{.15\textwidth}
		\centering
		\includegraphics[width=\linewidth]{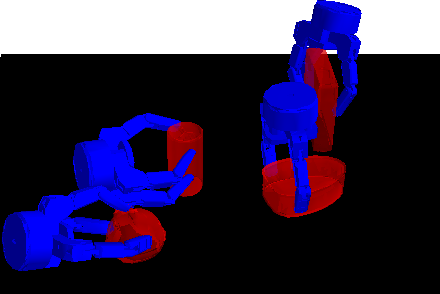}%
	\end{subfigure}\hspace{-0.3em}
	\begin{subfigure}[b]{.15\textwidth}
		\centering
		\includegraphics[width=\linewidth]{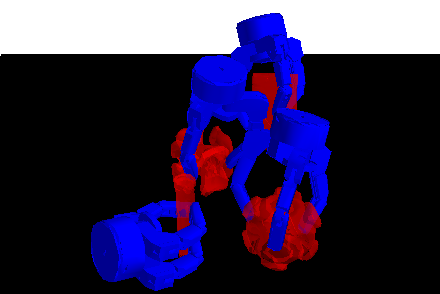}%
	\end{subfigure}
\\\hspace{-0.5em}
	\begin{subfigure}[b]{.15\textwidth}
		\centering
		\includegraphics[width=\linewidth]{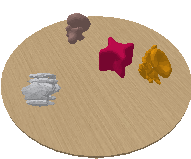}%
	\end{subfigure}\hspace{-0.3em}
	\begin{subfigure}[b]{.15\textwidth}
		\centering
		\includegraphics[width=\linewidth]{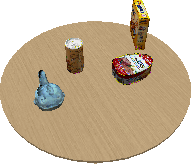}%
	\end{subfigure}\hspace{-0.3em}
	\begin{subfigure}[b]{.15\textwidth}
		\centering
		\includegraphics[width=\linewidth]{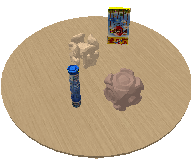}%
	\end{subfigure}
	\caption{\label{fig:grasps_on_sim_obj}Some example grasps shown in the top row proposed by \methodname{} in simulated scenes using the RGB input shown on the bottom row. The black background represents the table surface \bestcolor.}
    \vspace{-1em}
\end{figure}

The results are presented in \tabref{tb:sim_exp_summary} and \figref{fig:behaviour_over_clutter}. What is immediately
evident is that \methodname{} generates significantly higher quality grasps than both \multifin{} and \graspit{}. \methodname{} is also over 4 times faster than \graspit{}. As shown in
\figref{fig:epsilon_quality_over_clutter}, with increasing clutter the performance of \graspit{} deteriorates heavily much due to the optimization falls into local minimas between objects while the performance of \methodname{} and \multifin{} is constant. The main reason \multifin{} generates lower $\epsilon$-quality grasps than \methodname{} is because 27.25\% of its grasps are located below the table compared to only 1.25\% by \methodname. Some example grasps using \methodname{} are shown in \figref{fig:grasps_on_sim_obj}.

\figref{fig:finger_spread_histogram} shows a histogram over the finger
spread of the  \barrett{} produced by \methodname{}. Based on this figure, we can conclude that \methodname{}
produces diverse grasps, but seems to slightly favor grasps with no spread. One
reason zero-spread grasps, which are similar to parallel-jaw grasps, are
favored over others is because those grasps have smaller occupancy footprint
due to the kinematic structure of the \barrett{}.

In summary, the results from the simulation experiments show that \methodname{}
is able to quickly sample high-quality collision-free grasps in cluttered scenes.
Next, we will investigate its performance on real robotic hardware.

\subsection{Physical Grasping}
\label{exp:physical_grasping}

\begin{figure*}
	\centering
	\begin{subfigure}[b]{.13\textwidth}
		\centering
		\includegraphics[width=\linewidth]{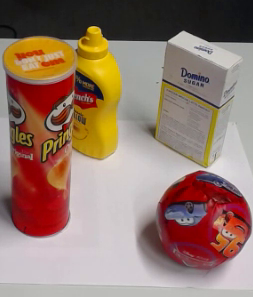}%
	\end{subfigure}
	\begin{subfigure}[b]{.158\textwidth}
		\centering
		\includegraphics[width=\linewidth]{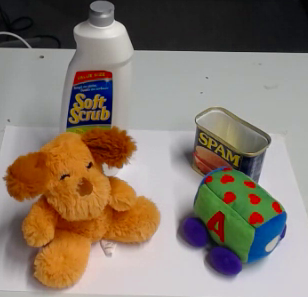}%
	\end{subfigure}
	\begin{subfigure}[b]{.13\textwidth}
		\centering
		\includegraphics[width=\linewidth]{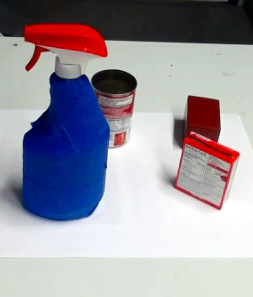}%
	\end{subfigure}
	\begin{subfigure}[b]{.13\textwidth}
		\centering
		\includegraphics[width=\linewidth]{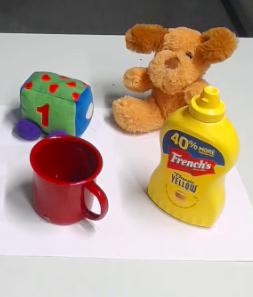}%
	\end{subfigure}
	\begin{subfigure}[b]{.13\textwidth}
		\centering
		\includegraphics[width=\linewidth]{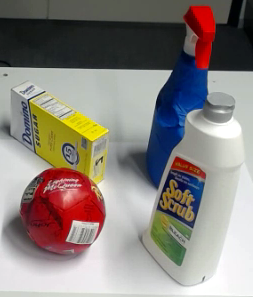}%
	\end{subfigure}
	\begin{subfigure}[b]{.13\textwidth}
		\centering
		\includegraphics[width=\linewidth]{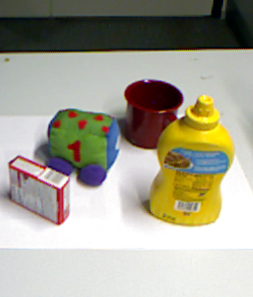}%
	\end{subfigure}
	\begin{subfigure}[b]{.13\textwidth}
		\centering
		\includegraphics[width=\linewidth]{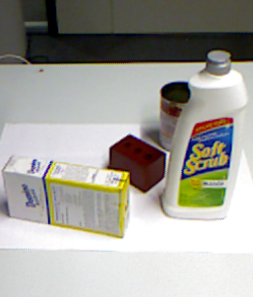}%
	\end{subfigure}
	\caption{\label{fig:test_scenes}Scenes used for testing}
	\vspace{-1em}
\end{figure*}

\begin{table}
    \centering
	\ra{1.3}\tbs{3}
	\caption{\label{tb:real_exp_summary}Real hardware experiment results}
    \begin{tabular}{@{}lccc@{}}
        \toprule
        & \multicolumn{1}{l}{GraspIt!} & \multicolumn{1}{l}{\multifin{}} & \multicolumn{1}{l}{\methodname{}} \\
        \midrule
        Average clearance rate (\%) $\uparrow$        & 60.71   & 57.15 & \textbf{71.43} \\
        Average grasp success rate (\%) $\uparrow$        & 32.72   & 34.90 & \textbf{40.00} \\
        Grasp Sampling for 10 grasps (sec.) $\downarrow$ & 40.40 & 10.70 & \textbf{8.10} \\
        \bottomrule
    \end{tabular}
	\vspace{-1.2em}
\end{table}

As a final experiment we studied the clearance rate of \methodname{}, \multifin{}, and \graspit{} on real robotic hardware. The clearance rate objective measures how many objects a grasping method can remove in a cluttered scene, given a predefined grasping budget. The setup consists of a \panda{} equipped with a \barrett{} and a \kinect{} camera to capture the input RGB-D images. The camera views the scene at a 55\textdegree{} viewing angle. For extrinsic calibration, we used an Aruco marker~\cite{garrido2014automatic}. We added the shape-completed objects to the planning scene to ensure collision-free motion planning.

As shown in \figref{fig:network} our method requires an input RGB image along with the individual object masks, and segmented \pcs{} of each object as seen by the depth image. To segment each object we first subtracted the background and the table and then extracted each segment with the Euclidean Cluster Extraction method in PCL\footnote{\href{https://pcl.readthedocs.io/en/latest/cluster\_extraction.html}{pcl.readthedocs.io/en/latest/cluster\_extraction.html}}. To generate the object masks, we transformed each of the \pc{}-segments to a binary depth-image and multiplied these with the input RGB image. The object masks were also used to generate grasps for \multifin{}. \figref{fig:network} shows examples of an input RGB image, object mask, and segmented \pc{}. Each segmented \pc{} was shape-completed into a mesh and the average of its vertices was used as the center of the object which was subsequently added to the translation produced by the Grasp Generation network. 

\methodname{}, \multifin{}, and \graspit{}  were benchmarked once on each of the seven different scenes shown in \figref{fig:test_scenes}. Each scene contained four objects of varying size, shape, and softness. The objects were manually placed to create scenes where objects occluded each other. We chose a grasping budget of eight grasps which is twice the number of objects in the scene, a heuristic that has been used in similar works~\cite{murali20206, lundell2020beyond}. A grasp was considered a success if the robot grasped the object and moved it to the start position without dropping it. If an object in the scene was accidentally moved or dropped, it was manually reset to its original pose.

The experimental results (\tabref{tb:real_exp_summary}) indicate that \methodname{} reaches a higher average clearance rate and grasp success rate than both \multifin{} and \graspit{}. Again, \methodname{} is significantly faster than \graspit{}, producing and evaluating grasps roughly 5 times as fast. The performance gain of \methodname{} compared to \multifin{} indicates that encoding information of the complete scene is useful for generating successful grasps in clutter, a finding that was also reported in \cite{sundermeyer2021contact}. For all methods the scene completion took, on average, 4 seconds. Both \methodname{} and \multifin{} were robust to noisy shape-completions and object masks.

 Examples of successful grasps generated with \methodname{} are shown in \figref{fig:successful_real_grasps}. These examples show that \methodname{} is able to produce a diverse set of grasp poses and configurations compatible with the clutter in the scene.

\subsection{Discussion}

Together, all results show that \methodname{} is able to generate high-quality collision-free grasps in cluttered simulated and real-world scenes. When comparing it to \graspit{}, \methodname{} achieves better performance, in a fraction of the time. The difference in performance is particularly significant in simulation, where \methodname{} achieves significantly higher $\epsilon$-quality than \graspit{} which explicitly optimizes grasps on such a metric. One potential reason for this difference is that the Finger Refinement layer gracefully refines the fingers to establish multiple contact points with the object resulting in higher quality grasps. This difference was not as evident in the real-world experiment, indicating that low $\epsilon$-qualities do not necessarily translate to unsuccessful grasps.

It is especially noteworthy that although \methodname{} is trained purely on simulated data it generalized seamlessly to real-world scenes without any fine-tuning. We hypothesize that this level of generalization is due to the grasps being originally generated in the object's reference frame, which correspond to the simulated data, and only later transformed to world coordinates by adding the center of the mesh to the pose. This effectively makes the network translation invariant and it can therefore focus on generating high-quality collision-free grasps no matter the scale.  

\begin{figure}%
	\centering
	\begin{subfigure}[b]{0.61\linewidth}
		\centering
		\includegraphics[width=.49\linewidth]{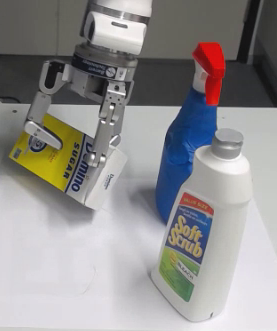}%
		\includegraphics[width=.49\linewidth]{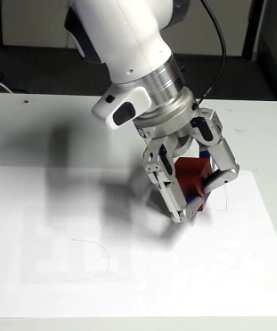}%
		\caption{\label{fig:successful_real_grasps}}
	\end{subfigure}
	\begin{subfigure}[b]{.30\linewidth}
		\centering
		\includegraphics[width=\linewidth]{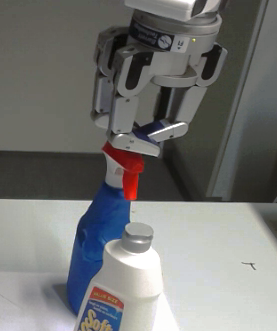}%
	    \caption{\label{fig:failed_real_grasps}}
	\end{subfigure}
	\caption{\label{fig:successful_and_failed_grasps}Some successful (a) and failed (b) grasps}
	\vspace{-1em}
\end{figure}

Nevertheless, \methodname{} also produced some catastrophically poor grasps. One example of such a grasp is shown in \figref{fig:failed_real_grasps}. Here the generated grasp was a precise pinch grasp where only the fingertips should touch the object, but due to inaccurate extrinsic calibration and low friction between the \barrett{} and the target object such grasps typically failed. In other failure cases, the gripper completely missed the object. One potential reason such grasps scored high is that they originally focused on\eg{} a corner of the object and due to imperfect extrinsic calibration the grasp simply missed the object. Both of these issues could be addressed by priming \methodname{} to produce more power type grasps.

Although \methodname{} is 4-5 times faster than \graspit{}, sampling 100 grasps still takes several seconds. The main reason for such a long sampling time stems from the expensive collision calculation in 
Eq. \eqref{eq:coll_loss} which accounts for more than 85\% of the total computational time. Nevertheless, sub-second sampling is still possible for 10 or fewer grasps.  
\section{Conclusions and future work}
\label{sec:conclusions}

We presented \methodname{}, a generative deep network that produces high-quality 6-\ac{dof} dexterous grasps in structured clutter in a matter of seconds. Grasping in clutter with dexterous multi-finger hands is especially difficult as the added occupancy footprint of such grippers makes it non-trivial to generate collision-free grasps. \methodname{} overcomes such difficulties by combining scene-completion, scene-encoding, and a fully differentiable forward kinematics layer to generate dexterous collision-free grasps. We compared \methodname{} to \multifin{} and \graspit{} in both simulation and real-world grasping and the results show that \methodname{} produces higher quality grasps than both of these and is over 4-5 times faster than \graspit{}.

Despite the impressive results, there is still room for improvements. In terms of sampling speed, \methodname{} could be sped up significantly if instead of doing explicit collision checking, it would rather use a separate network for collision detection as in~\cite{murali20206}. Another limiting factor is that the high-quality grasps produced by \methodname{} do not always translate to successful real grasps. Instead of explicitly calculating the quality metric, a critique network to score multi-finger grasps is probably a good solution but this is still a more open problem than on parallel-jaw grasps~\cite{mousavian20196,mahler2017dex}.

In conclusion, the work presented here shows that we can quickly generate high-quality collision-free multi-finger grasps in clutter. This, in turn, enables the use of multi-finger grasps in practical applications which has so far been limited due to the excessive computation time needed by other methods.

\section*{Acknowledgment}
We gratefully acknowledge the support of NVIDIA Corporation with the donation of the Titan Xp GPU used for this research.

\bibliographystyle{IEEEtran}
\bibliography{refs}

\begin{thebibliography}{10}
\providecommand{\url}[1]{#1}
\csname url@samestyle\endcsname
\providecommand{\newblock}{\relax}
\providecommand{\bibinfo}[2]{#2}
\providecommand{\BIBentrySTDinterwordspacing}{\spaceskip=0pt\relax}
\providecommand{\BIBentryALTinterwordstretchfactor}{4}
\providecommand{\BIBentryALTinterwordspacing}{\spaceskip=\fontdimen2\font plus
\BIBentryALTinterwordstretchfactor\fontdimen3\font minus
  \fontdimen4\font\relax}
\providecommand{\BIBforeignlanguage}[2]{{%
\expandafter\ifx\csname l@#1\endcsname\relax
\typeout{** WARNING: IEEEtran.bst: No hyphenation pattern has been}%
\typeout{** loaded for the language `#1'. Using the pattern for}%
\typeout{** the default language instead.}%
\else
\language=\csname l@#1\endcsname
\fi
#2}}
\providecommand{\BIBdecl}{\relax}
\BIBdecl

\bibitem{mahler2017dex}
J.~Mahler, J.~Liang, S.~Niyaz, M.~Laskey, R.~Doan, X.~Liu, J.~A. Ojea, and
  K.~Goldberg, ``Dex-net 2.0: Deep learning to plan robust grasps with
  synthetic point clouds and analytic grasp metrics,'' \emph{arXiv:1703.09312},
  2017.

\bibitem{morrison2018closing}
D.~Morrison, P.~Corke, and J.~Leitner, ``Closing the loop for robotic grasping:
  A real-time, generative grasp synthesis approach,'' \emph{arXiv:1804.05172},
  2018.

\bibitem{levine_learning_2016}
S.~Levine, P.~Pastor, A.~Krizhevsky, and D.~Quillen, ``Learning {Hand}-{Eye}
  {Coordination} for {Robotic} {Grasping} with {Deep} {Learning} and
  {Large}-{Scale} {Data} {Collection},'' \emph{arXiv:1603.02199 [cs]}, 2016.

\bibitem{murali20206}
A.~Murali, A.~Mousavian, C.~Eppner, C.~Paxton, and D.~Fox, ``6-dof grasping for
  target-driven object manipulation in clutter,'' in \emph{2020 IEEE
  International Conference on Robotics and Automation (ICRA)}.\hskip 1em plus
  0.5em minus 0.4em\relax IEEE, 2020, pp. 6232--6238.

\bibitem{kokic2017affordance}
M.~Kokic, J.~A. Stork, J.~A. Haustein, and D.~Kragic, ``Affordance detection
  for task-specific grasping using deep learning,'' in \emph{2017 IEEE-RAS 17th
  International Conference on Humanoid Robotics (Humanoids)}.\hskip 1em plus
  0.5em minus 0.4em\relax IEEE, 2017, pp. 91--98.

\bibitem{miller2004graspit}
A.~T. {Miller} and P.~K. {Allen}, ``Graspit! a versatile simulator for robotic
  grasping,'' \emph{IEEE Robotics Automation Magazine}, vol.~11, no.~4, pp.
  110--122, 2004.

\bibitem{wu2020generative}
B.~Wu, I.~Akinola, A.~Gupta, F.~Xu, J.~Varley, D.~Watkins-Valls, and P.~K.
  Allen, ``Generative attention learning: a "general" framework for
  high-performance multi-fingered grasping in clutter,'' \emph{Autonomous
  Robots}, vol.~44, pp. 971--990, 2020.

\bibitem{lundell2020multi}
J.~Lundell, E.~Corona, T.~N. Le, F.~Verdoja, P.~Weinzaepfel, G.~Rogez,
  F.~Moreno-Noguer, and V.~Kyrki, ``Multi-fingan: Generative coarse-to-fine
  sampling of multi-finger grasps,'' \emph{arXiv:2012.09696}, 2020.

\bibitem{satish2019policy}
V.~Satish, J.~Mahler, and K.~Goldberg, ``On-policy dataset synthesis for
  learning robot grasping policies using fully convolutional deep networks,''
  \emph{IEEE Robotics and Automation Letters}, vol.~4, no.~2, pp. 1357--1364,
  2019.

\bibitem{lundell2020beyond}
J.~Lundell, F.~Verdoja, and V.~Kyrki, ``Beyond top-grasps through scene
  completion,'' in \emph{2020 IEEE International Conference on Robotics and
  Automation (ICRA)}.\hskip 1em plus 0.5em minus 0.4em\relax IEEE, 2020, pp.
  545--551.

\bibitem{ten2017grasp}
A.~ten Pas, M.~Gualtieri, K.~Saenko, and R.~Platt, ``Grasp pose detection in
  point clouds,'' \emph{The International Journal of Robotics Research},
  vol.~36, no. 13-14, pp. 1455--1473, 2017.

\bibitem{gualtieri2016high}
M.~Gualtieri, A.~Ten~Pas, K.~Saenko, and R.~Platt, ``High precision grasp pose
  detection in dense clutter,'' in \emph{2016 IEEE/RSJ International Conference
  on Intelligent Robots and Systems (IROS)}.\hskip 1em plus 0.5em minus
  0.4em\relax IEEE, 2016, pp. 598--605.

\bibitem{sundermeyer2021contact}
M.~Sundermeyer, A.~Mousavian, R.~Triebel, and D.~Fox, ``Contact-graspnet:
  Efficient 6-dof grasp generation in cluttered scenes,''
  \emph{arXiv:2103.14127}, 2021.

\bibitem{varley2015generating}
J.~Varley, J.~Weisz, J.~Weiss, and P.~Allen, ``Generating multi-fingered
  robotic grasps via deep learning,'' in \emph{2015 IEEE/RSJ International
  Conference on Intelligent Robots and Systems (IROS)}.\hskip 1em plus 0.5em
  minus 0.4em\relax IEEE, 2015, pp. 4415--4420.

\bibitem{berenson2008grasp}
D.~Berenson and S.~S. Srinivasa, ``Grasp synthesis in cluttered environments
  for dexterous hands,'' in \emph{8th IEEE-RAS International Conference on
  Humanoid Robots}.\hskip 1em plus 0.5em minus 0.4em\relax IEEE, 2008, pp.
  189--196.

\bibitem{sahbani2012overview}
A.~Sahbani, S.~El-Khoury, and P.~Bidaud, ``An overview of 3d object grasp
  synthesis algorithms,'' \emph{Robotics and Autonomous Systems}, vol.~60,
  no.~3, pp. 326--336, 2012.

\bibitem{bohg_data-driven_2014}
J.~Bohg, A.~Morales, T.~Asfour, and D.~Kragic, ``Data-{Driven} {Grasp}
  {Synthesis} - {A} {Survey},'' \emph{IEEE Transactions on Robotics}, 2014.

\bibitem{miller_automatic_2003}
A.~T. Miller, S.~Knoop, H.~I. Christensen, and P.~K. Allen, ``Automatic grasp
  planning using shape primitives,'' in \emph{2003 {IEEE} {International}
  {Conference} on {Robotics} and {Automation}}, vol.~2, 2003, pp. 1824--1829.

\bibitem{ciocarlie2009hand}
M.~T. Ciocarlie and P.~K. Allen, ``Hand posture subspaces for dexterous robotic
  grasping,'' \emph{The International Journal of Robotics Research}, vol.~28,
  no.~7, pp. 851--867, 2009.

\bibitem{borst2003grasping}
C.~Borst, M.~Fischer, and G.~Hirzinger, ``Grasping the dice by dicing the
  grasp,'' in \emph{Proceedings 2003 IEEE/RSJ International Conference on
  Intelligent Robots and Systems (IROS 2003)(Cat. No. 03CH37453)},
  vol.~4.\hskip 1em plus 0.5em minus 0.4em\relax IEEE, 2003, pp. 3692--3697.

\bibitem{goldfeder2007grasp}
C.~Goldfeder, P.~K. Allen, C.~Lackner, and R.~Pelossof, ``Grasp planning via
  decomposition trees,'' in \emph{Proceedings 2007 IEEE International
  Conference on Robotics and Automation}.\hskip 1em plus 0.5em minus
  0.4em\relax IEEE, 2007, pp. 4679--4684.

\bibitem{pelossof2004svm}
R.~Pelossof, A.~Miller, P.~Allen, and T.~Jebara, ``An svm learning approach to
  robotic grasping,'' in \emph{IEEE International Conference on Robotics and
  Automation, 2004. Proceedings. ICRA'04. 2004}, vol.~4.\hskip 1em plus 0.5em
  minus 0.4em\relax IEEE, 2004, pp. 3512--3518.

\bibitem{varley_shape_2017}
J.~Varley, C.~DeChant, A.~Richardson, J.~Ruales, and P.~Allen, ``Shape
  completion enabled robotic grasping,'' in \emph{{IEEE}/{RSJ} {International}
  {Conference} on {Intelligent} {Robots} and {Systems}}, 2017, pp. 2442--2447.

\bibitem{lundell2019robust}
J.~Lundell, F.~Verdoja, and V.~Kyrki, ``Robust {Grasp} {Planning} {Over}
  {Uncertain} {Shape} {Completions},'' in \emph{2019 {IEEE}/{RSJ}
  {International} {Conference} on {Intelligent} {Robots} and {Systems}
  ({IROS})}.\hskip 1em plus 0.5em minus 0.4em\relax Macau, China: IEEE, Nov.
  2019.

\bibitem{watkins-valls_multi-modal_2018}
D.~Watkins-Valls, J.~Varley, and P.~Allen, ``Multi-{Modal} {Geometric}
  {Learning} for {Grasping} and {Manipulation},'' \emph{arXiv:1803.07671},
  2018.

\bibitem{agnew2020amodal}
W.~Agnew, C.~Xie, A.~Walsman, O.~Murad, C.~Wang, P.~Domingos, and S.~Srinivasa,
  ``Amodal 3d reconstruction for robotic manipulation via stability and
  connectivity,'' \emph{arXiv:2009.13146}, 2020.

\bibitem{shao2020unigrasp}
L.~Shao, F.~Ferreira, M.~Jorda, V.~Nambiar, J.~Luo, E.~Solowjow, J.~A. Ojea,
  O.~Khatib, and J.~Bohg, ``Unigrasp: Learning a unified model to grasp with
  multifingered robotic hands,'' \emph{IEEE Robotics and Automation Letters},
  vol.~5, no.~2, pp. 2286--2293, 2020.

\bibitem{aktas2019deep}
U.~R. Aktas, C.~Zhao, M.~Kopicki, A.~Leonardis, and J.~L. Wyatt, ``Deep
  dexterous grasping of novel objects from a single view,''
  \emph{arXiv:1908.04293}, 2019.

\bibitem{lu2020active}
Q.~Lu, M.~Van~der Merwe, and T.~Hermans, ``Multi-fingered active grasp
  learning,'' \emph{arXiv:2006.05264}, 2020.

\bibitem{corona2020ganhand}
E.~Corona, A.~Pumarola, G.~Alenya, F.~Moreno-Noguer, and G.~Rogez, ``Ganhand:
  Predicting human grasp affordances in multi-object scenes,'' in
  \emph{Proceedings of the IEEE/CVF Conference on Computer Vision and Pattern
  Recognition}, 2020, pp. 5031--5041.

\bibitem{lorensen_marching_1987}
\BIBentryALTinterwordspacing
W.~E. Lorensen and H.~E. Cline, ``Marching {Cubes}: {A} {High} {Resolution} 3d
  {Surface} {Construction} {Algorithm},'' in \emph{Proceedings of the 14th
  {Annual} {Conference} on {Computer} {Graphics} and {Interactive}
  {Techniques}}, ser. {SIGGRAPH} '87.\hskip 1em plus 0.5em minus 0.4em\relax
  New York, NY, USA: ACM, 1987, pp. 163--169. [Online]. Available:
  \url{http://doi.acm.org/10.1145/37401.37422}
\BIBentrySTDinterwordspacing

\bibitem{martin2017wasserstein}
S.~Martin~Arjovsky and L.~Bottou, ``Wasserstein generative adversarial
  networks,'' in \emph{Proceedings of the 34 th International Conference on
  Machine Learning, Sydney, Australia}, 2017, pp. 214--223.

\bibitem{gulrajani2017improved}
I.~Gulrajani, F.~Ahmed, M.~Arjovsky, V.~Dumoulin, and A.~C. Courville,
  ``Improved training of wasserstein gans,'' in \emph{Advances in neural
  information processing systems}, 2017, pp. 5767--5777.

\bibitem{feix2015grasp}
T.~Feix, J.~Romero, H.-B. Schmiedmayer, A.~M. Dollar, and D.~Kragic, ``The
  grasp taxonomy of human grasp types,'' \emph{IEEE Transactions on
  human-machine systems}, vol.~46, no.~1, pp. 66--77, 2015.

\bibitem{mousavian20196}
A.~Mousavian, C.~Eppner, and D.~Fox, ``6-dof graspnet: Variational grasp
  generation for object manipulation,'' in \emph{Proceedings of the IEEE
  International Conference on Computer Vision}, 2019, pp. 2901--2910.

\bibitem{coumans2019}
E.~Coumans and Y.~Bai, ``Pybullet,'' \url{http://pybullet.org}, 2016.

\bibitem{calli2015ycb}
B.~Calli, A.~Singh, A.~Walsman, S.~Srinivasa, P.~Abbeel, and A.~M. Dollar,
  ``The ycb object and model set: Towards common benchmarks for manipulation
  research,'' in \emph{2015 international conference on advanced robotics
  (ICAR)}.\hskip 1em plus 0.5em minus 0.4em\relax IEEE, 2015, pp. 510--517.

\bibitem{morrison2020egad}
D.~Morrison, P.~Corke, and J.~Leitner, ``Egad! an evolved grasping analysis
  dataset for diversity and reproducibility in robotic manipulation,''
  \emph{IEEE Robotics and Automation Letters}, vol.~5, no.~3, pp. 4368--4375,
  2020.

\bibitem{kasper2012kit}
A.~Kasper, Z.~Xue, and R.~Dillmann, ``The kit object models database: An object
  model database for object recognition, localization and manipulation in
  service robotics,'' \emph{The International Journal of Robotics Research},
  vol.~31, no.~8, pp. 927--934, 2012.

\bibitem{miller1999examples}
A.~T. Miller and P.~K. Allen, ``Examples of 3d grasp quality computations,'' in
  \emph{Proceedings 1999 IEEE International Conference on Robotics and
  Automation}, vol.~2.\hskip 1em plus 0.5em minus 0.4em\relax IEEE, 1999, pp.
  1240--1246.

\bibitem{garrido2014automatic}
S.~Garrido-Jurado, R.~Mu{\~n}oz-Salinas, F.~J. Madrid-Cuevas, and M.~J.
  Mar{\'\i}n-Jim{\'e}nez, ``Automatic generation and detection of highly
  reliable fiducial markers under occlusion,'' \emph{Pattern Recognition},
  vol.~47, no.~6, pp. 2280--2292, 2014.

\end{thebibliography}

\end{document}